%% file: accv2018cameraready.tex
\documentclass[runningheads]{llncs}
\usepackage{graphicx}
\usepackage{amsmath,amssymb} 
\usepackage{color}
\usepackage{multicol}
\usepackage{mathtools}
\usepackage[ruled]{algorithm2e}
\usepackage{wrapfig,lipsum,booktabs}

\def\eg{\emph{e.g.}}
\def\etal{\emph{et al.}}

\definecolor{brown}{rgb}{0.65, 0.16, 0.16}

\begin{document}

\title{Universal Bounding Box Regression\\and Its Applications} 
\titlerunning{Universal Bounding Box Regression} 


\author{Seungkwan Lee \and
Suha Kwak \and
Minsu Cho }
%

\authorrunning{S. Lee et al.} 


\institute{Dept. of Computer Science and Engineering, POSTECH, Korea\\ 
\email{\{seungkwan,suha.kwak,mscho\}@postech.ac.kr}}

\maketitle

\input{_0_abstract.tex}

\input{_1_introduction.tex}

\input{_2_method.tex}

\input{_3_experiment.tex}

\input{_4_conclusion.tex}

\paragraph{\bf Acknowledgements:}
This research was supported by Samsung Research and also by Basic Science Research Program through the National Research Foundation of Korea funded by the Ministry of Science, ICT (NRF-2018R1A5A1060031, NRF-2017R1E1A1A01077999).

\bibliographystyle{splncs04}
\bibliography{egbib}

\end{document}

%% file: _0_abstract.tex
\begin{abstract}
Bounding-box regression is a popular technique to refine or predict localization boxes in recent object detection approaches. Typically, bounding-box regressors are trained to regress from either region proposals or fixed anchor boxes to nearby bounding boxes of a pre-defined target object classes. This paper investigates whether the technique is generalizable to unseen classes and is transferable to other tasks beyond supervised object detection. 
To this end, we propose a class-agnostic and anchor-free box regressor, dubbed {\em Universal Bounding-Box Regressor} (UBBR), which predicts a bounding box of the nearest object from any given box. Trained on a relatively small set of annotated images, UBBR successfully generalizes to unseen classes, and can be used to improve localization in many vision problems. We demonstrate its effectivenss on weakly supervised object detection and object discovery.

\keywords{Bounding box regression  \and Transfer learning \and Weakly-supervised object detection}
\end{abstract}

%% file: _1_introduction.tex

\section{Introduction}
The recent advances in object detection have been driven mainly by the development of Deep Neural Networks (DNNs)~\cite{girshick2014rich,fast_rcnn,ren2015faster,liu2016ssd,redmon2016you,YOLOv2,huang2017speed}.
Especially, one crucial component that allows DNNs to localize object bounding boxes precisely and flexibly is the Bounding Box Regressor (BBR) originally proposed in~\cite{girshick2014rich}.
As a part of object detection networks, BBR refines off-the-shelf object proposals~\cite{girshick2014rich,fast_rcnn} or anchor boxes with fixed positions and aspect ratios~\cite{ren2015faster,liu2016ssd,YOLOv2} so that the refined ones localize nearby objects more accurately. 
For this purpose, BBRs are tightly coupled with other components of object detection networks, and trained to localize predefined object classes better.
That is, they have been developed typically for supervised object detection where ground-truth bounding boxes for target classes are given.

This paper studies BBR in a direction different from the conventional one.
Specifically, we propose a BBR model that is class-agnostic, even well generalizable to unseen classes, and transferable to multiple diverse tasks demanding accurate bounding box localization; we call such a model \emph{Universal Bounding Box Regressor} (UBBR).
UBBR takes an image and any arbitrary bounding boxes, and refines the boxes so that they enclose their nearest objects tightly, regardless of their classes.
The model with such a simple functionality can have a great impact on many applications since it is universal in terms of both object classes and tasks.
An example of the applications is weakly supervised object detection where box annotations for target object classes are not given.
In this setting, object bounding boxes tend to be badly localized due to the limited supervision~\cite{Bilen16,kantorov2016contextlocnet,oicr}, and UBBR can help to improve the performance by refining the localization results.
In this case, UBBR can be considered as a knowledge transfer machine for bounding box localization. 
Also, UBBR can be used to generate object box proposals.
Given boxes uniformly and densely sampled from image space, UBBR transforms them to approximate the boxes of their nearest objects, and the results are bounding boxes clustered around true object boxes.
In this case, UBBR can be considered as learning-based object proposal methods~\cite{zitnick2014edge,Pinheiro2015,Pinheiro2016}.

This paper introduces a DNN architecture for UBBR and its training strategy.
Our UBBR has a form of Convolutional Neural Networks (CNN), trained with randomly generated input boxes.
It successfully generalizes to unseen classes, and can be used to improve localization in various computer vision problems, especially when bounding box supervision is absent.
We demonstrate its effectivenss on weakly supervised object detection, object proposal generation, and object discovery.
Main contribution of this paper is three-fold:
\begin{itemize}
   \item We present a simple yet effective UBBR based on CNN, which is versatile and easily generalizable to unseen classes. We also present a training strategy to learn such a universal model.
   \item A single UBBR network achieves, or help to achieve, competitive performance in three different  applications: weakly supervised object detection, object proposals, and object discovery.
   \item We provide an in-depth empirical analysis for demonstrating the generalizability of our UBBR for unseen classes. 
   
\end{itemize}

The rest of this paper is organized as follows.
Section~\ref{sec:related} overviews previous approaches relevant to UBBR, and Section~\ref{sec:method} presents technical details of UBBR and a strategy for training it.
UBBR is then evaluated on three different localization tasks in Section~\ref{sec:exp}, and we conclude in Section~\ref{sec:conclusion} with brief remarks.


\section{Related Work}
\label{sec:related}

\paragraph{\bf Conventional BBR in Object Detection:}
BBR has been widely incorporated into DNNs for object detection~\cite{girshick2014rich,fast_rcnn,ren2015faster,liu2016ssd,redmon2016you,YOLOv2} for precise localization of object bounding boxes.
Initially it was designed as a post-processing step to refine off-the-shelf object proposals boxes~\cite{girshick2014rich,fast_rcnn}.
Recently, it directly estimates bounding boxes of nearby objects from each cell of an image grid~\cite{redmon2016you}, 
or aims to transform a fixed set of anchor boxes to cover ground-truth object boxes accurately~\cite{ren2015faster,liu2016ssd,YOLOv2}.
Here the anchor boxes, also known as default boxes, are pre-defined bounding boxes that are sampled on a regular grid with a few selected scales and aspect ratios~\cite{ren2015faster,redmon2016you,liu2016ssd} or estimated from ground-truth object boxes of training data~\cite{YOLOv2}.
Thus those BBRs are trained to be well harmonized with other components of object detection networks, and are dependent on a few pre-defined object classes and characteristics of anchor boxes. 
On the other hand, our UBBR is designed and trained to be class-agonstic, transferable to unseen classes, and free from anchor boxes.
These properties of UBBR allow us to apply it to multiple diverse applications demanding accurate bounding box localization, beyond the conventional object detection.

\paragraph{\bf Object Proposal:}
Our UBBR is also closely related to object proposals since it naturally generates  accurate object candidate boxes given uniformly sampled boxes as inputs.
Well-known early approaches to object proposal are unsupervised techniques~\cite{uijlings2013selective,manen2013prime}.
Motivated by the fact that typically an object box include a whole image segment rather than a part of it,
they draw bounding boxes encompassing image segments obtained by hierarchical image segmentation methods.
Since there is no supervision for object location and image segmentation results often fail to preserve object boundary, the unsupervised techniques are limited in terms of recall and localization accuracy.
Supervised approaches for object proposals have been actively studied as well, and exhibited substantially better performance.
Before the era of deep learning, there have been proposed object proposal techniques generating object candidate boxes~\cite{zitnick2014edge} and masks~\cite{arbelaez2014multiscale}, which are trained with object boundary annotations.
Recently, Pinheiro~\etal~\cite{Pinheiro2015,Pinheiro2016} introduce DNNs for generating and refining class-agnostic object candidate masks.

Learning-based proposals, including ours, require strong supervision in training. One may ask, if such bounding box annotations are given, why not directly learning an object detector instead of proposals? 
We would like to argue that the learning-based proposals are still valuable if they are class-agnostic, well generalizable to unseen classes, and universally applied to various applications.
Note that existing datasets provide a huge amount of readily available annotations, especially for bounding boxes; there is no reason to avoid them when localizing objects of unseen classes in the context of transfer learning.

\paragraph{\bf Transfer Learning for Visual Recognition:}
Oquab~\etal~\cite{Oquab2014} demonstrated that low-level layers of a CNN trained for a large-scale image classification can be transferred to classification in different domains or even different visual recognition tasks.
Since that, transferring low-level image representation has been a common technique to avoid overfitting in various visual recognition tasks like object detection~\cite{girshick2014rich,fast_rcnn,ren2015faster,liu2016ssd,redmon2016you,YOLOv2,huang2017speed} and semantic segmentation~\cite{Fcn,vggnet,deeplab_v2}.
While these approaches focus on transferring low-level image representation between different tasks, UBBR is to transfer the knowledge about \emph{how to draw bounding boxes to enclose an object}. 
In that sense, UBBR also has a connection to TransferNet~\cite{transfernet}, which transfers the segmentation knowledge to object classes whose segmentation annotations are not available.

%% file: _2_method.tex

\section{Universal Bounding Box Regressor}
\label{sec:method}
\subsection{Architecture}
\begin{figure}[t]
\centering
\includegraphics[width=\textwidth]{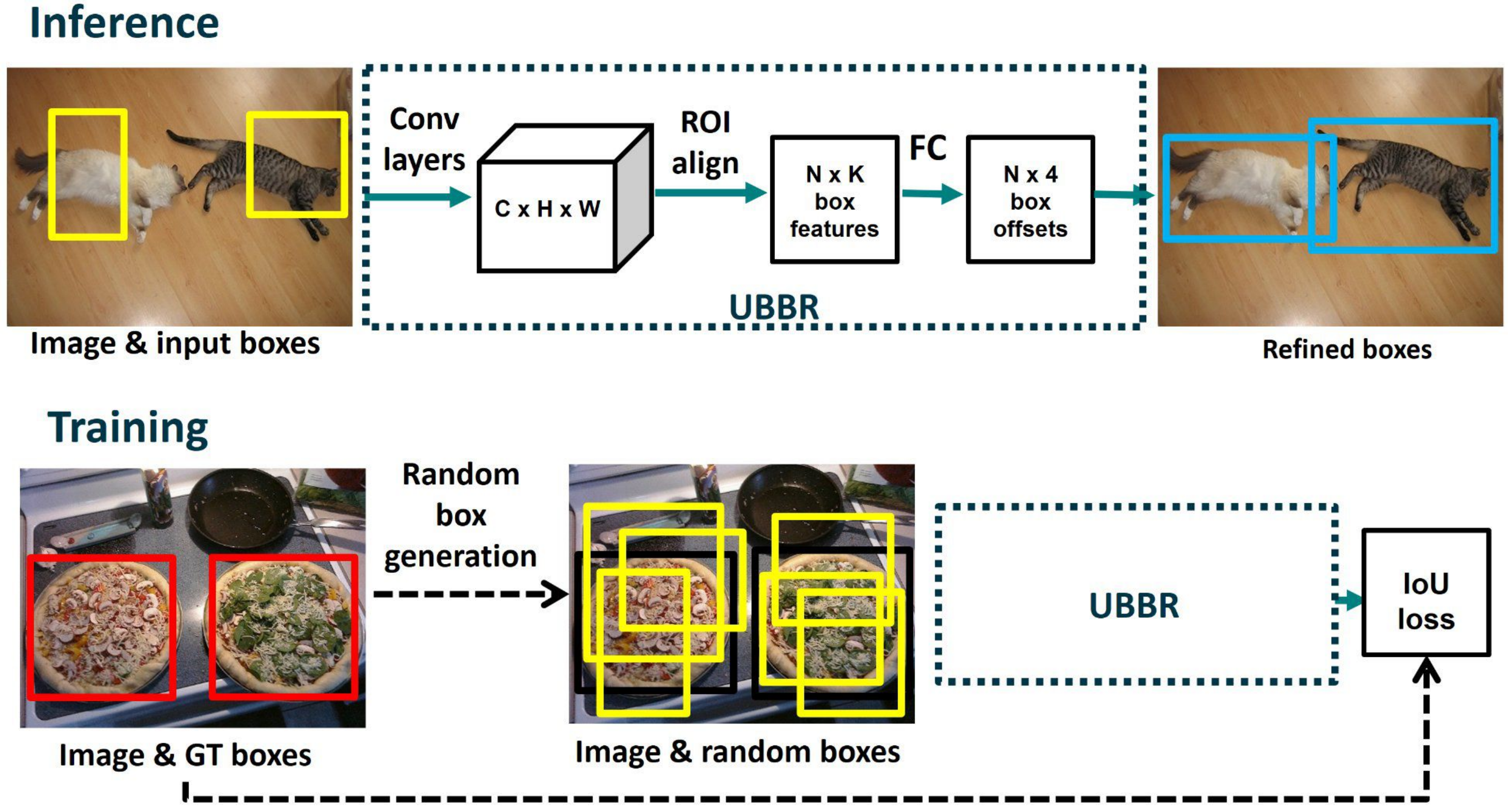}
\caption{
Illustration of UBBR's architecture. 
In inference time, the network takes an image with roughly localized bounding boxes and refine them so that they tightly enclose nearby objects.
$N$ is the number of input boxes and $K$ is the dimensionality of box features.
In training time, the network takes bounding boxes randomly generated around ground-truth boxes, and is learned to transform each input box so that Intersection-over-Union between the box and its nearest ground-truth is maximized.
}
\label{fig:archi}

\end{figure}

The architecture of UBBR is similar with conventional object detectors (\eg, Fast R-CNN \cite{fast_rcnn}) which consist of convolutional layers for feature representation, a region pooling layer for extracting region-wise features, and fully-connected layers for box classification and regression. 
Figure~\ref{fig:archi} illustrates training and inference stages of the UBBR network. The architecture first computes a feature map of an input image with the convolutional layers, and a feature vector of a fixed length is extracted for each input box through the RoI-Align layer~\cite{he2017mask}.
Each of the extracted box features is then processed by 3 fully-connected layers to compute a 4-D real vector indicating the offset between the corresponding box and its nearest object.
Note that UBBR is designed to use input boxes with arbitrary shapes and object classes unlike those of most conventional object detection networks~\cite{ren2015faster,fast_rcnn}. 
Hence, the UBBR network is trained in a anchor-free and class-agnostic manner as will be described in the following.

\subsection{Training}

\paragraph{\bf Dataset:}
Since UBBR predicts object boxes, it demands images with ground-truth object boxes during training, and any existing datasets for object detection can meet the need.
Note that since UBBR is class-agnostic, class labels of the box annotations are disregarded in our case.

\paragraph{\bf Random Box Generation:}
\begin{figure}[t]
\centering
\includegraphics[width=\textwidth]{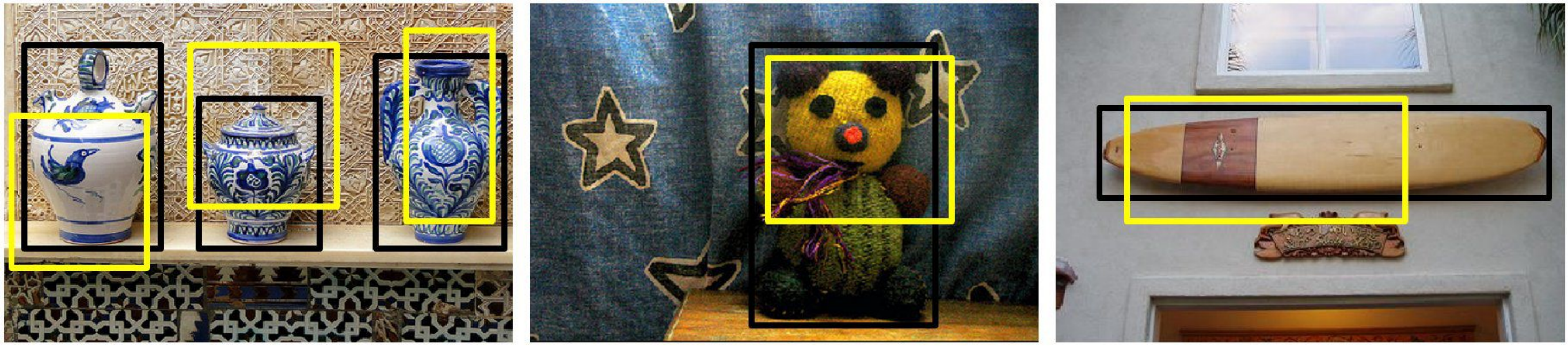}
\caption{Example of randomly generated bounding boxes for training UBBR. 
Black boxes are ground-truths and yellow ones are randomly generated boxes.}
\label{fig:rand}
\end{figure}

UBBR takes as its inputs not only image but also (roughly localized) boxes that will be transformed to enclose nearby objects tightly.
Thus, each training image has to be served together with such boxes.
Furthermore, the boxes fed to the network during training should be diverse for universality of UBBR, but at the same time, have to be overlapped with at least one ground-truth to some extent so that UBBR can observe enough evidences about target object.
To this end, in training time we generate input bounding boxes by applying random transformations to ground-truth boxes.

Let $g = [x_g, y_g, w_g, h_g]^\top$ denote a ground-truth box represented by its center coordinate $(x_g, y_g)$, width $w_g$, and height $h_g$.
Transformation parameters for the four values are sampled from uniform distributions independently as follows:
\begin{equation}
\begin{split}
&t_x \sim \mathcal{U}(-\alpha,\; \alpha),\\
&t_y \sim \mathcal{U}(-\alpha,\; \alpha),\\
&t_w \sim \mathcal{U}(\ln{1 - \beta},\; \ln{1 + \beta}),\\
&t_h \sim \mathcal{U}(\ln{1 - \beta},\; \ln{1 + \beta}).
\end{split}
\end{equation}
Then a random input box $b=[x_b, y_b, w_b, h_b]^\top $ is obtained by applying the sampled transformation to $g$:
\begin{equation}
\begin{split}
&x_b = x_g + t_x \cdot w_g,\\
&y_b = y_g + t_y \cdot h_g,\\
&w_b = w_g \cdot \exp(t_w),\\
&h_b = h_g \cdot \exp(t_h).
\end{split}
\end{equation}
Also, if Intersection-over-Union (IoU) between $b$ and $g$ is less than a pre-defined threshold $t$, we simply discard $b$ during training.
$\alpha$ and $\beta$ are empirically set to 0.35 and 0.5 respectively. 
The effect of $\alpha$, $\beta$, and $t$ on the performance of UBBR is analyzed in the next section.
Figure~\ref{fig:rand} shows examples of random box generation.

\paragraph{\bf Loss Function:}
For the regression criterion, IoU loss~\cite{yu2016unitbox} is employed instead of conventional ones like $L_2$ and smooth $L_1$ losses.
The drawback of the conventional losses in bounding box regression is that the bounding box transformation parameters $(t_x, t_y, t_w, t_h)$ are optimized independently~\cite{yu2016unitbox} although they are in fact highly inter-correlated.
IoU loss has been proposed to address this issue, and we observed in our experiments that IoU loss allows training more stable and leads to better performance when compared to smooth $L_1$ loss.

\begin{algorithm}[t!] \label{alg:IoULoss}
    \SetKwInOut{Input}{Input}
    \SetKwInOut{Output}{Output}

    \Input{Two bounding boxes $u = [x_u, y_u, w_u, h_u]^\top$, and $v = [x_v, y_v, w_v, h_v]^\top$}
    \Output{loss $\mathcal{L}$}
    \caption{IoU loss}
    
  \DontPrintSemicolon
  \SetKwFunction{FMain}{IoU-loss}
  \SetKwProg{Fn}{Function}{:}{}
  \Fn{\FMain{$u$, $v$}}{
    $A_u = w_u \cdot h_u$ \\
    $A_v = w_v \cdot h_v$ \\
    $I_w = \min(x_u + 0.5 \cdot w_u,\; x_v + 0.5 \cdot w_v) - \max(x_u - 0.5 \cdot w_u,\; x_v - 0.5 \cdot w_v)$ \\
    $I_h = \min(y_u + 0.5 \cdot h_u,\; y_v + 0.5 \cdot h_v) - \max(y_u - 0.5 \cdot h_u,\; y_v - 0.5 \cdot h_v)$ \\
    $I_w = \max(I_w,\; 0)$\\ 
    $I_h = \max(I_w,\; 0)$\\
    $I = I_w \cdot I_h$ \\
    $U = A_u + A_v - I$ \\
    $IoU = \frac{I}{U}$ \\
    $\mathcal{L} = -\ln{(IoU + \epsilon)}$ \\
        \KwRet\ $\mathcal{L}$
  }
\end{algorithm}

The procedure for computing IoU loss between two bounding boxes is described in Algorithm~\ref{alg:IoULoss}, where
$A_u$ and $A_v$ are the areas of $u$ and $v$, and $I_w$ and $I_h$ means the width and height of their intersection area.
Note that we add a tiny constant $\epsilon$ to IoU value before taking logarithm for numerical stability. 
The image-level loss is then defined as the average of box-wise regression losses as follows:
\begin{equation}
\begin{split}
    &  L_{\textrm IoU} = \frac{1}{N} \sum_{n=1}^{N} \textrm{IoU-loss}\Big(f\big(b_n, \textrm{UBBR}(b_n)\big), g_n \Big), 
\end{split}
\end{equation}
where $b_n$ is an input box and $g_n$ is the ground-truth bounding box that is best overlapped with $b_n$ in terms of IoU metric. 
Also, UBBR($b_n$) is the offsets predicted by UBBR and $f$ is the transformation function that refines $b_n$ with the predicted offset parameters.

%% file: _3_experiment.tex

\section{Experiment}
\label{sec:exp}

In this section, we first describe implementation details,
then demonstrate the effectiveness of our approach empirically in three tasks: weakly supervised object detection, object proposal, and object discovery.

\subsection{Datasets}
To demonstrate transferability of UBBR, we carefully define source and target domains.
Basically, we employ COCO 2017 \cite{Mscoco} as source and PASCAL VOC \cite{Pascalvoc} as target.
Then all images containing the 20 PASCAL VOC object categories are removed from the COCO 2017.
As a result, there remain 21,413 training images and 900 validation images of 60 object categories in the source domain dataset.
Note that we train a single UBBR with the above dataset, and apply the model to all applications without task-specific finetuning.

\subsection{Implementation Details}
The training is carried out using stochastic gradient decent with momentum and weight decay.
The momentum and weight decay multiplier are set to 0.9 and 0.0005, respectively.
The learning rate initially starts from $10^{-3}$ and is divided by 10 when the validation loss stop improving.
We stop the training when the learning rate become $10^{-6}$.
In all experiments, we employ ResNet101 \cite{resnet} (upto conv4) pre-trained on ImageNet as backbone convolutional layers.
The fully-connected layers are composed of three linear layers with ReLU activations.
The weight parameters of fully connected layers are randomly initialized from zero-mean Gaussian distributions with standard deviation 0.001, and their biases are initialized to 0.
For both training and testing, input images are rescaled using bilinear interpolation such that its shorter side becomes 600 pixels.
We generate 50 random bounding boxes for each ground-truth object.

\subsection{Weakly Supervised Object Detection}
\begin{table}[t]
\resizebox{\linewidth}{!}{
\begin{tabular}{ |c| c c c c c c c c c c c c c c c c c c c c|c|} 
\hline
Method & aer & bik & brd & boa & btl & bus & car & cat & cha & cow & tbl & dog & hrs & mbk & prs & plt & shp & sfa & trn & tv & mAP \\ 
\hline
\hline
OICR-paper & 58.0 & 62.4 & 31.1 & 19.4 & 13.0 & 65.1 & 62.2 & 28.4 & 24.8 & 44.7 & 30.6 & 25.3 & 37.8 & 65.5 & 15.7 & 24.1 & 41.7 & 46.9 & 64.3 & 62.6 & 41.2 \\
OICR-ours(baseline) & 61.2 &64.6 &41.3 &24.1 &10.4 &65.7 &62.3 &32.6 &23.1 &48.0 &35.3 &29.3 &43.8 &63.9 &14.1 &24.0 &41.3 &50.5 &61.1 &61.1 & 42.9 \\
\hline
OICR + UBBR-iou($t$=0.5) &  66.0  &58.0&   50.8  &31.3&   17.9& 71.1& 66.6& 47.7& 26.2& 59.1& 40.6& 40.6& 54.8& 63.4& 23.3& 25.3& 51.1& 57.7& 68.0& 66.3 & 49.3\\
OICR + UBBR-iou($t$=0.3) &  66.3& 56.9& 53.9& 32.4& 22.4& 71.3& 67.3& 53.4& 25.5& 60.0 &40.4 &47.0&  61.6& 64.0 &28.3 &25.5&   51.4& 61.1& 67.7& 67.6& 51.2\\
\hline
OICR + UBBR-sl1($t$=0.5) &  65.7& 57.0 &49.9& 30.5& 18.7& 69.5 &66.2 &45.6& 25.6& 58.9& 40.9& 39.9& 56.9& 65.2& 21.7& 25.5& 50.7& 56.8& 67.7 &65.9& 48.9\\

OICR + UBBR-sl1($t$=0.3) &  65.2& 52.1 &53.7& 30.3& 22.2& 71.4& 66.8& 52.6& 23.6& 60.5& 37.5& 47.1& 61.9& 63.7& 27.3& 24.4& 51.4& 58.5& 69.3& 66.2& 50.3\\

\hline
\end{tabular}
}
\\
\caption{
Average precision (\emph{IoU} $>$ 0.5) for weakly supervised object detection on PASCAL VOC 2007 test set. For baseline model, we train OICR using published code and extract detection results from it. We refer to this model as OICR-ours. $t$ is \emph{IoU} threshold for random box generation. The models trained with smooth L1 and IoU losses are denoted by UBBR-sl1 and UBBR-iou, respectively.
}
\label{table:wsd1}

\resizebox{\linewidth}{!}{
\begin{tabular}{ |c| c c c c c c c c c c c c c c c c c c c c|c|} 
\hline
Method & aer & bik & brd & boa & btl & bus & car & cat & cha & cow & tbl & dog & hrs & mbk & prs & plt & shp & sfa & trn & tv & mAP \\ 
\hline
\hline

OICR + UBBR($t$=0.5) 1 iter & 66.0  &58.0&   50.8  &31.3&   17.9& 71.1& 66.6& 47.7& 26.2& 59.1& 40.6& 40.6& 54.8& 63.4& 23.3& 25.3& 51.1& 57.7& 68.0& 66.3 & 49.3\\
OICR + UBBR($t$=0.5) 2 iter & 63.9& 50.4& 53.8& 32.1& 23.3& 73.3& 66.9& 52.6& 25.9& 64.4& 38.1& 47.3& 58.6& 62.7& 27.0& 23.2& 55.2& 60.2& 68.6& 66.3& 50.7  \\
OICR + UBBR($t$=0.5) 3 iter & 59.9& 48.3& 55.4& 34.9& 24.6& 73.8& 66.8& 60.7& 25.5& 63.5& 35.1& 51.4& 59.5  &62.9&   31.0& 22.0& 56.1& 60.8& 69.6& 66.2& 51.4\\
\hline
OICR + UBBR($t$=0.3) 1 iter & 66.3& 56.9& 53.9& 32.4& 22.4& 71.3& 67.3& 53.4& 25.5& 60.0&   40.4& 47.0& 61.6& 64.0 &   28.3& 25.5& 51.4& 61.1& 67.7& 67.6& 51.2\\
OICR + UBBR($t$=0.3) 2 iter & 63.2& 47.2& 55.2& 33.8& 27.4& 71.7& 67.5& 67.9& 24.0& 62.6& 33.1& 58.6& 63.2& 63.4& 35.7& 19.1& 52.9& 58.3& 67.8& 63.9& 51.8  \\
OICR + UBBR($t$=0.3) 3 iter & 59.7  &44.8 &54.0 &  36.1& 29.3& 72.1& 67.4& 70.7& 23.5& 63.8& 31.5& 61.5& 63.7& 61.9& 37.9& 15.4& 55.1& 57.4& 69.9& 63.6& 52.0\\

\hline
\end{tabular}
}
\\
\caption{
Performance improvement of iterative refinement.
}
\label{table:wsd3}
\end{table}
\begin{figure}[!t]
\centering
\includegraphics[width=\textwidth]{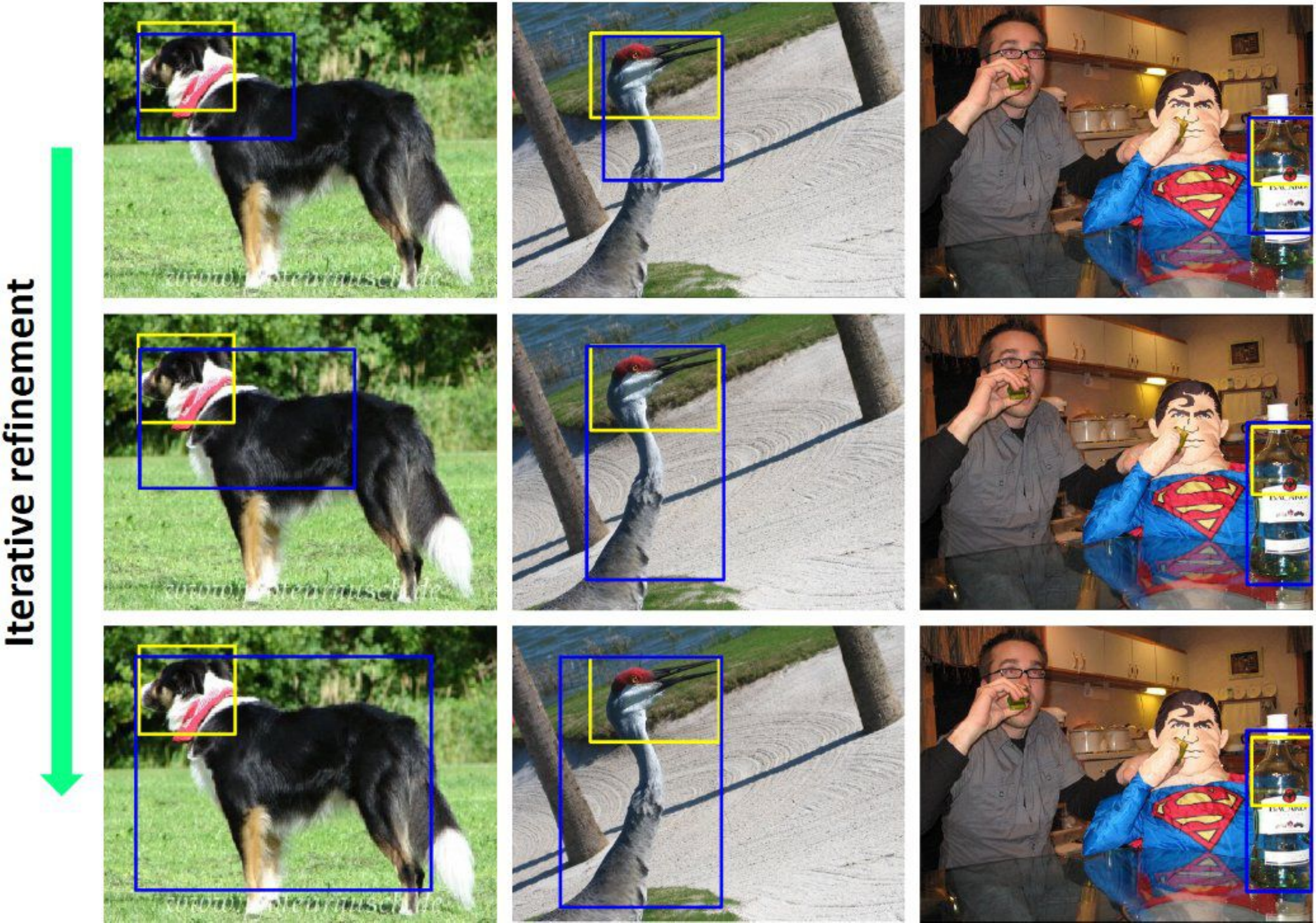}
\caption{
Qualitative results of (OICR + UBBR) on PASCAL VOC 2007 test set. Yellow boxes are detection results of OICR and blue boxes are refined bounding boxes. From top to bottom, each row is the result of 1, 2, and 3 iterative refinement respectively.
}
\label{fig:wsd_qual}
\end{figure}
To demonstrate the effectiveness of UBBR, we apply our model as a post-processing module of weakly supervised object detection.
The goal of weakly supervised object detection is to learn object detectors only with image-level class labels as supervision.
Due to the significantly limited supervision, models in this category often fail to localize the entire body of target object but cover only a discriminative part of it. 
Thus, UBBR can help to improve localization by refining bounding boxes estimated by weakly supervised object detection model.
This setting also can be considered as transfer learning for weakly supervised object detection, where UBBR transfer the bounding box knowledge of source domain to target domain.

We use OICR \cite{oicr} as a baseline model for weakly supervised object detection, and apply UBBR to the output of OICR.
The quantitative analysis of the performance on PASCAL VOC 2007 is summarized in Table~\ref{table:wsd1}, in which one can see that UBBR improves the object localization quality substantially.
We also validate the effect of the threshold $t$ by applying UBBR models learned with two different values of $t$.
In general, the model with a smaller $t$ performs better than that with a larger $t$ since UBBR is able to learn from more various and challenging box localization examples by decreasing $t$ during training.
Also, we report the performance of the models learned with conventional smooth $L_1$ loss.
Figure~\ref{fig:wsd_qual} presents qualitative results of our approach.

Besides the above straightforward application of UBBR, we further explore ways to better utilize UBBR and provide more detailed analysis on its various aspects in the context of weakly supervised object detection as follows.
\begin{figure}[t]
\centering
\includegraphics[width=\textwidth]{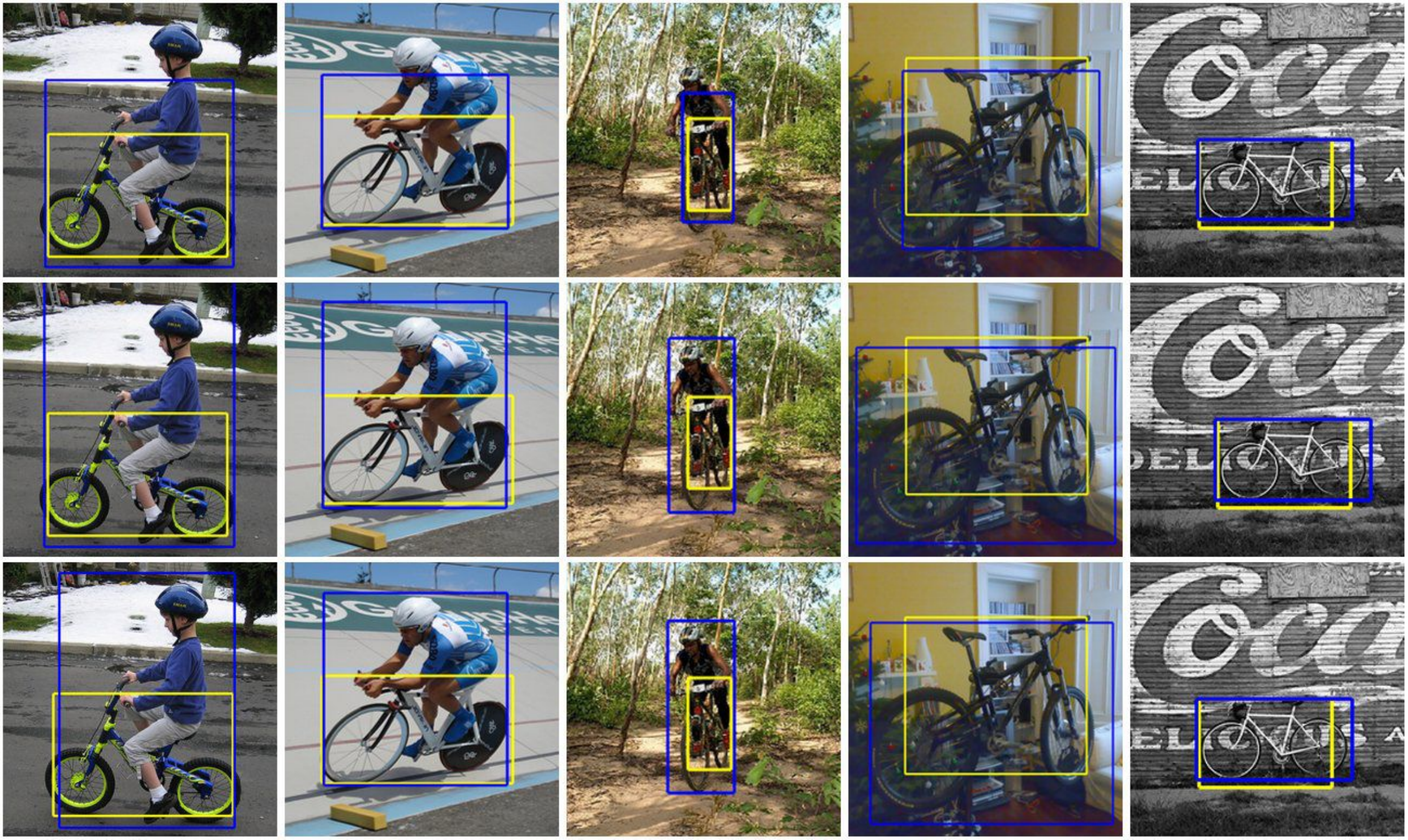}
\caption{
Box refinement examples of bike class. Yellow boxes are detection results of OICR and blue boxes are refined bounding boxes. From top to bottom, each row is the result of 1, 2, and 3 iterative refinement respectively. Left three examples are failure cases, and right two examples are successful cases.
}
\label{fig:fail_case}
\end{figure}
\paragraph{\bf Iterative Refinement:}

UBBR also can be applied multiple times iteratively so that localization is progressively improved. 
That is, for each iteration, bounding boxes refined in previous step are fed into the network again.
Through this strategy, we can obtain better localization results. 
It is important to note that, for efficiency of overall procedure, we reuse the convolutional feature map of the backbone network.
As can be seen in Table~\ref{table:wsd3}, we can further improve the localization performance by iterative refinement, and the effect was consistent up to the third iterations.

\paragraph{\bf Limitation:}

As Table~\ref{table:wsd1} shows, the quality of refined localization of {\it bike} class is worse than baseline.
Furthermore, the iterative refinement makes the quality even worse as shown in Table~\ref{table:wsd3}.
This means UBBR rather degrades localization of {\it bike} class, and we found that it is because of a side effect of the class-agnostic nature of UBBR. 
Figure~\ref{fig:fail_case} shows box refinement examples of {\it bike} class.
Left three examples are failure cases, and right two examples are successful cases.
Most of failure cases of {\it bike} class occur when there is a person riding the bike.
Because UBBR predicts class-agnostic bounding box, it does not distinguish {\it bike} and {\it person} and recognizes them as a single object in the examples.
As illustrated in two rightmost columns, when there is no person on the bike, it successfully localizes the bikes.
\begin{table}[t]
\resizebox{\linewidth}{!}{
\begin{tabular}{ |c| c c c c c c c c c c c c c c c c c c c c|c|} 
\hline
Method & aer & bik & brd & boa & btl & bus & car & cat & cha & cow & tbl & dog & hrs & mbk & prs & plt & shp & sfa & trn & tv & mAP \\ 
\hline
\hline
OICR-ours(baseline) & 61.2 &64.6 &41.3 &24.1 &10.4 &65.7 &62.3 &32.6 &23.1 &48.0 &35.3 &29.3 &43.8 &63.9 &14.1 &24.0 &41.3 &50.5 &61.1 &61.1 & 42.9 \\
\hline

COCO-21($t$=0.5) & 65.2&  66.6  &46.0&  31.2& 19.2& 67.0&   65.7& 36.6& 26.3& 51.5& 35.7& 32.9& 49.5& 66.0&   16.1& 25.0 &43.6&   56.2& 62.0&   65.1& 46.4 \\
COCO-21($t$=0.3) & 63.8&  67.7& 48.9& 30.7& 21.5& 67.2& 66.2& 37.8& 25.2& 52.1& 38.9& 34.8& 48.9& 65.2& 18.8  &23.9 &38.2 &57.5&   62.4& 65.4& 46.8 \\
\hline
COCO-40($t$=0.5) & 65.4&  65.9& 50.6& 30.7& 18.8& 66.9& 65.7& 44.4& 26.2& 55.1& 38.8& 36.8& 54.1& 66.4& 17.8  &24.8 &46.9&   56.1& 63.7& 63.9& 48.0 \\
COCO-40($t$=0.3) &65.2 &63.7&   50.4& 30.4& 23.3& 69.8& 66.0 &46.3&   25.7& 56.9& 41.8& 42.7& 56.6& 65.3& 21.3& 24.1  &46.4&   59.8& 62.2& 63.9& 49.1 \\
\hline
COCO-60($t$=0.5) &  66.0  &58.0&   50.8  &31.3&   17.9& 71.1& 66.6& 47.7& 26.2& 59.1& 40.6& 40.6& 54.8& 63.4& 23.3& 25.3& 51.1& 57.7& 68.0& 66.3 & 49.3\\
COCO-60($t$=0.3) &  66.3& 56.9& 53.9& 32.4& 22.4& 71.3& 67.3& 53.4& 25.5& 60.0 &40.4 &47.0&  61.6& 64.0 &28.3 &25.5&   51.4& 61.1& 67.7& 67.6& 51.2\\
\hline
COCO-full($t$=0.5) &  66.4&	64.5&	51.4&	34.2&	19.7&	72.0&	67.0&	47.7&	26.3&	56.9&	41.4&	38.7&	57.0&	65.5&	26.8&	26.4&	50.7&	56.5	&70.8	&64.4 & 50.2\\
COCO-full($t$=0.3) &  67.6&	63.9&	54.1&	33.0&	24.1&	72.7&	69.0&	53.4	&26.1&	59.1&	42.1&	47.7&	63.1&	65.6&	38.4&	28.1&	51.9&	60.0&	70.7&	66.6&	52.9\\
\hline

\end{tabular}
}
\\
\caption{
Average precision (\emph{IoU} $>$ 0.5) for weakly supervised object detection on PASCAL VOC 2007 test set. COCO-60 is our main dataset excluding 20 categories from original COCO 2017 dataset. COCO-21 and COCO-40 are more reduced datasets which contain 21 and 40 categories respectively. COCO-full is the original COCO 2017 train set which contains 80 categories.
}
\label{table:wsd4}
\centering
\resizebox{4cm}{!}{
\begin{tabular}{|c|c|c|c|}
\hline
& $\beta$=0.35 & $\beta$=0.5 & $\beta$=0.65  \\ 
\hline
$\alpha$=0.25 &  49.0  &   51.2  & 51.7\\ 
\hline
$\alpha$=0.35 &  48.9  &    51.2 &  52.0\\
\hline
$\alpha$=0.45 &   49.0 &    50.7 &51.8  \\
\hline
\end{tabular}
}
\caption{
Effect of box generation parameters $\alpha$ and $\beta$ on the performance of weakly-supervised object detection. $\alpha = 0.35$ and $\beta = 0.5$ are used in all other experiments.}
\label{table:box_ablation}
\end{table}

\paragraph{\bf Generalizability:}
The previous experiments already validated that our approach is generalizable to unseen object classes of the target domain.
To further demonstrate the generalizability, we analyze the performance of UBBR models trained with even a smaller number of object classes. 
To this end, we build two additional training sets by reducing the number of object classes.
COCO-40 is composed of 40 categories excluding animal, accessory, electronic, and appliance classes from the original training data.
Also, COCO-21 consists of 21 classes and is obtained by further excluding furniture, indoor, and food classes from COCO-40.
The original training dataset is denoted by COCO-60.
Moreover, to eliminate the effect of dataset size, we make the sizes of COCO-40 and COCO-21 identical to that of COCO-60 by randomly sampling 21,413 images containing at least one object belonging to the categories of interest.

We report the performance of UBBRs learned with COCO-40 and COCO-21 in Table~\ref{table:wsd4}. 
Although the models trained with these datasets perform worse due to lack of diversity in their training data, they still improve localization performance substantially.
An interesting observation is that they improve localization of animals although their training datasets do not include animal classes.
The results indicate that UBBR can be generalizable to unseen and unfamiliar classes well.
We also report the performance of UBBR models learned with full COCO 2017 train set, which is denoted by COCO-full and contains all PASCAL VOC classes.
It is natural that UBBR trained with COCO-full outperforms the others, but their differences in performance are marginal. 

\paragraph{\bf Box generation parameters:}
The box generation parameter $\alpha$ and $\beta$ are chosen empirically to generate diverse and sufficiently overlapped boxes. 
Table~\ref{table:box_ablation} shows how these parameters affect the performance of weakly-supervised object detection when $t$ is 0.3. 
As shown in the table, the performance is not very sensitive to both parameters.
In all other experiments, $\alpha = 0.35$ and $\beta = 0.5$ are used.
Note that we did not optimize those parameters using the evaluation results.
\begin{figure}[!t]
\centering
\includegraphics[width=\textwidth]{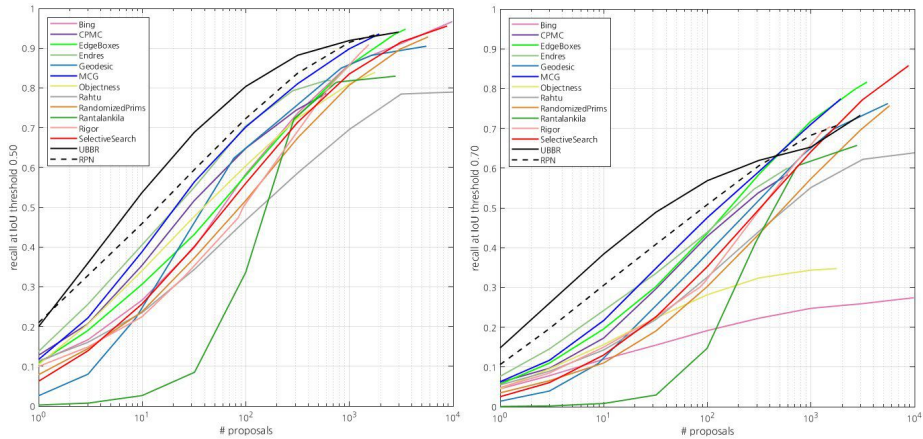}
\caption{
Recall of box proposals on the PASCAL VOC 2007 test set. 
(\emph{left}) recall@IOU=0.5. (\emph{right}) recall@IOU=0.7.
}
\label{fig:prop_example}
\end{figure}

\subsection{Object Proposals}

\begin{figure}[t]
\centering
\includegraphics[width=\textwidth]{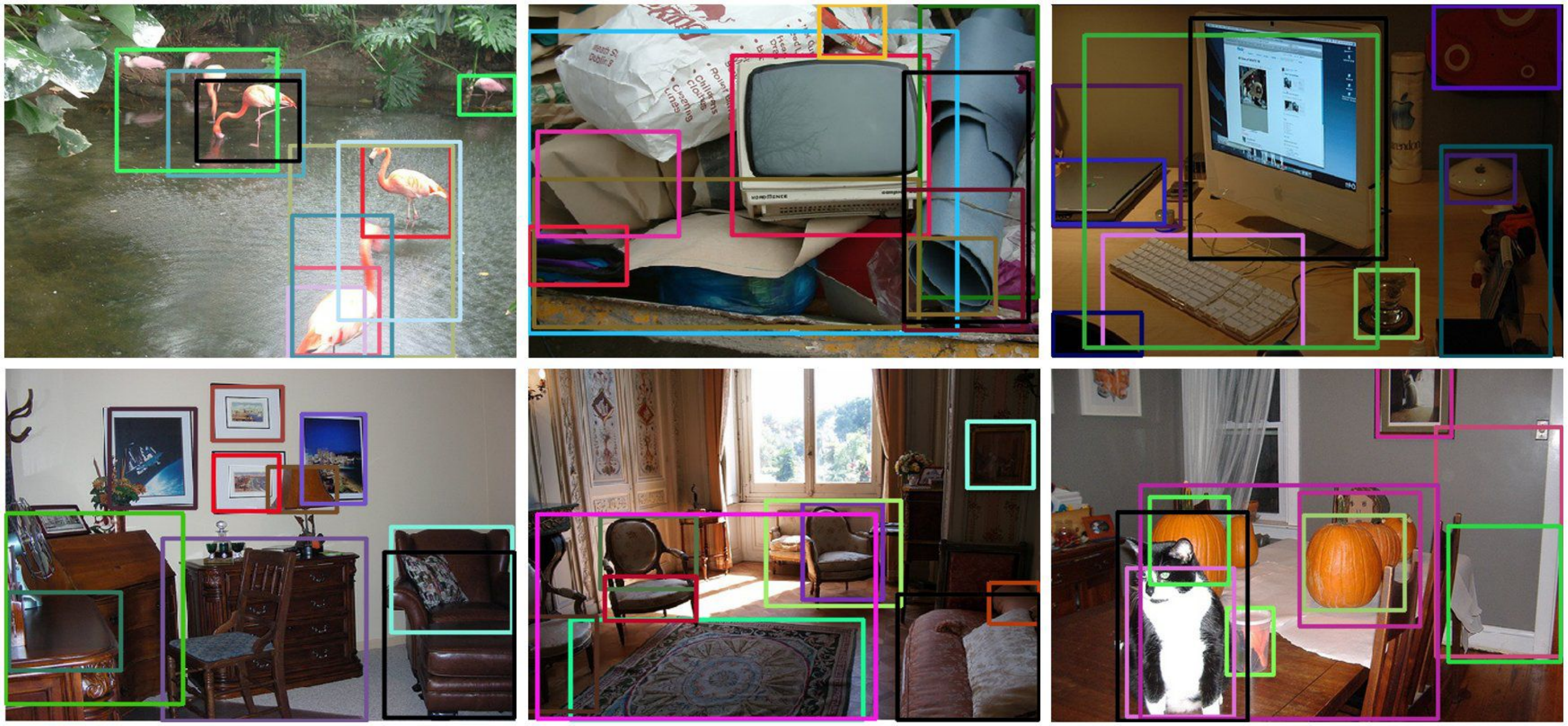}
\caption{
Visualization of top-10 region proposals generated by the proposed method.
}
\label{fig:prop_performance}
\end{figure}

For the second application, we employ UBBR as a region proposal generator. 
Similarly to RPN \cite{ren2015faster}, we generate seed bounding boxes of various scale and aspect ratio and locate them in image uniformly.
We feed them into UBBR so that each seed bounding box encloses its nearest object.
To select object proposals from the refined bounding boxes, we assign score $s_n$ to each bounding box $b_n$.
In assumption that the refined bounding boxes will be concentrated around real objects, 
$s_n$ is initially set to the number of adjacent bounding boxes whose \emph{IoU} with $b_n$ is greater than 0.7.
After that, we apply non-maximum suppression (NMS) with \emph{IoU} threshold 0.6.
In NMS procedure, instead of removing adjacent bounding boxes, we divide their scores by 10, which is similar to Soft-NMS \cite{soft_nms}.
In Figure~\ref{fig:prop_example}, performance of proposals generated by our method are quantified and compared with popular proposal techniques~\cite{uijlings2013selective,manen2013prime,zitnick2014edge,arbelaez2014multiscale,cheng2014bing,carreira2012cpmc,endres2014category,krahenbuhl2014geodesic,alexe2012measuring,rahtu2011learning,rantalankila2014generating,humayun2014rigor}.
The performance of UBBR clearly outperforms previous methods in comparison. 
Note that unlike many other methods (except SelectiveSearch~\cite{uijlings2013selective}), UBBR does not use any images from PASCAL object classes for training.  
We also evaluate RPN~\cite{ren2015faster} in the same transfer learning scenario with ours, where we train RPN with COCO-60 dataset and evaluate it on PASCAL VOC dataset. 
Note that we use the same backbone network for both of RPN and UBBR. 
As shown in Figure~\ref{fig:prop_example}, UBBR outperforms RPN in particular with a tighter IOU criterion.
Note that the x axis of the figure starts from recall at $10^0$ proposal rather than $10^1$ proposals.
Figure~\ref{fig:prop_performance} presents qualitative examples of object proposals obtained by our method.

\begin{table}[!t]
\resizebox{\linewidth}{!}{
\begin{tabular}{ |c| c c c c c c c c c c c c c c c c c c c c|c|} 
\hline
Method & aer & bik & brd & boa & btl & bus & car & cat & cha & cow & tbl & dog & hrs & mbk & prs & plt & shp & sfa & trn & tv &  Avg\\ 
\hline
\hline

Cho~\etal~\cite{cho2015unsupervised} &50.3& 42.8& 30.0& 18.5& 4.0& 62.3& 64.5& 42.5& 8.6& 49.0& 12.2& 44.0& 64.1 &57.2& 15.3& 9.4& 30.9& 34.0& 61.6& 31.5 &36.6\\
Li \etal~\cite{li2016image} &73.1& 45.0 &43.4& 27.7& 6.8 &53.3& 58.3 &45.0& 6.2 &48.0 &14.3 &47.3& 69.4& 66.8 &24.3& 12.8 &51.5 &25.5& 65.2& 16.8& 40.0\\
\hline
Ours &47.9  &18.9&   63.1 &   39.7& 10.2& 62.3 &   69.3& 61.0 &   27.0& 79.0 &   24.5 &   67.9 &   79.1 &   49.7& 28.6& 12.8& 79.4& 40.6& 61.6& 28.4& 47.6  \\

\hline
\end{tabular}
}
\\
\caption{
Object discovery accuracy in CorLoc on PASCAL VOC 2007 trainval set.
}
\label{table:discovery}

\end{table}
\subsection{Object Discovery}

For the last application, we choose the task of object discovery that aims at localizing objects from images. Since most of previous methods consider localization of a single foreground object per image, the object discovery can be viewed as an extreme case of object proposal generation where only top-1 proposals are used for evaluation. 
The correct localization (CorLoc) metric is an evaluation metric widely used in related work \cite{joulin2014efficient,cho2015unsupervised,li2016image}, and defined as the percentage of images correctly localized according to the PASCAL criterion: $\frac{area(b_p \cap b_{gt})}{
area(b_p \cup b_{gt})} > 0.5$, where $b_p$ is the predicted box and $b_{gt}$ is the ground-truth box.
For evaluation on the PASCAL VOC 2007 dataset, we follow  to use all images in PASCAL VOC 2007 trainval set discarding images which only contain `difficult' or `truncated' objects. We report the performance in Table \ref{table:discovery}. The performance of UBBR significantly outperforms the previous approaches to object discovery~\cite{cho2015unsupervised,li2016image}, which implies that generic object information can be effectively learned by UBBR and transferred to the task of object discovery.

%% file: _4_conclusion.tex

\section{Conclusion}
\label{sec:conclusion}

We have studied the bounding box regression in a novel and interesting direction.
Unlike those commonly embedded in recent object detection networks, our model is class-agnostic and free from manually defined anchor boxes. 
These properties allow our model to be universal, well generalizable to unseen classes, and transferable to multiple diverse tasks demanding accurate bounding box localization.
Such advantages of our model have been verified empirically in various tasks including weakly supervised object detection, object proposal, and object discovery.